\documentclass{article}
\usepackage[utf8]{inputenc} % Force LaTeX to interpret bytes as UTF-8

\usepackage{iclr2025_conference,times, graphicx, float, booktabs, hyperref, listings, xcolor, url}

\lstdefinelanguage{json}{
    basicstyle=\ttfamily\small\color{black}, % default font style and color
    numbers=left,
    numberstyle=\tiny\color{gray},
    stepnumber=1,
    numbersep=5pt,
    showstringspaces=false,
    breaklines=true,
    frame=single,
    framerule=0.5pt,
    rulecolor=\color{black},
    backgroundcolor=\color{white},
    literate=
     *{:}{{{\color{black}:}}}{1}
      {,}{{{\color{black},}}}{1}
      {"}{{{\color{black}"}}}{1},
    morestring=[b]",
    stringstyle=\color{red}, % makes strings red
    identifierstyle=\color{black}, % keys black
    commentstyle=\color{gray}
}

\title{AI for Monitoring and Classifying Data Used in Research Literature}

\author{
Rafael Macalaba$^{\dagger}$ \& Aivin V. Solatorio$^{*}$ \\
Development Data Group \\
Office of the World Bank Group Chief Statistician \\
The World Bank \\
1818 H Street N.W., \\
Washington, 20433 \\
District of Columbia, USA \\
\texttt{\{rmacalaba, asolatorio\}@worldbank.org}
}

\iclrfinalcopy % Uncomment for camera-ready (non-anonymous)

\begin{document}

\maketitle

\begingroup
\renewcommand\thefootnote{\fnsymbol{footnote}}
\footnotetext[1]{GitHub/HF: \href{https://github.com/avsolatorio}{\texttt{@avsolatorio}}, \href{mailto:avsolatorio@gmail.com}{avsolatorio@gmail.com}}
\footnotetext[2]{GitHub: \href{https://github.com/rafmacalaba}{\texttt{@rafmacalaba}}, \href{mailto:rafael.macalaba@yahoo.com}{rafael.macalaba@yahoo.com}}
\endgroup

\begin{abstract}
While platforms like Google Scholar and Semantic Scholar track citations for academic papers, no comparable infrastructure exists for monitoring dataset usage in research literature, leaving the landscape of data use largely opaque. Addressing this gap is critical for transparency, reproducibility, and monitoring of impact, yet progress is hindered by inconsistent citation practices, scarce labeled data, and ambiguous references to datasets in the wild. Traditional NLP approaches struggle with these challenges, motivating the shift toward more adaptive, semantically rich models. Building on prior work using LLMs for data mention detection and synthetic data for bootstrapping training, this paper presents an updated methodology for scalable dataset monitoring. We introduce a multitask GLiNER-based framework that jointly performs dataset mention extraction, relation identification, and usage-context classification. To address label scarcity, the pipeline leverages synthetic data generation to produce training examples and LLM-based revalidation to filter incorrect mentions and enforce labeling consistency, together improving reliability, coverage, and output consistency across the training pipeline. This work advances the development of open-source tools for monitoring data use in research literature, contributing to the broader goal of generalizable, unconstrained dataset citation tracking.
\end{abstract}

\section{Introduction}

Datasets are essential to empirical research, underpinning analysis, validation, and policy development \citep{mooney2012data, silvello2018citing}.  
However, tracking how datasets are cited and used across research literature remains challenging due to inconsistent citation practices and limited structured metadata \citep{buneman2021linked, piwowar2013data}. Manual efforts to extract dataset mentions are infeasible at scale, motivating recent applications of artificial intelligence (AI) and natural language processing (NLP) for automated detection \citep{potok2022automated, heddes2021automatic, hussain2025extracting}.  

Our previous work introduced a large language model (LLM) framework that combined synthetic data generation and LLM-based revalidation to enhance dataset-mention detection and labeling consistency \citep{solatorio2025ai}. While the approach achieved strong recall and effectively expanded low-resource training data, maintaining factual accuracy in model outputs remained challenging—general-purpose instruction-tuned models such as Phi-3 occasionally hallucinated or paraphrased dataset names instead of reproducing them verbatim. These observations motivated the development of a more specialized and controlled architecture designed for precise, structured extraction of dataset mentions and their attributes.

This paper introduces an updated methodology centered on a multitask GLiNER-based architecture \citep{zaratianna2023gliner}. The model unifies dataset-mention extraction, relation identification (e.g., producer, acronym, geography, year), and usage-context classification within a single training pipeline. Retaining the strengths of the original framework—synthetic-data generation and LLM-based revalidation—it simplifies model design and improves structured extraction accuracy.

Evaluated on the original annotated corpus and new document collections, the multitask GLiNER framework demonstrates consistent gains over previous baselines, including GLiNER-large-v2.1 \citep{zaratianna2023gliner} and NuExtract \citep{cripwell2024nuextract}. By extending dataset-use monitoring into an integrated extraction and classification framework, this update supports more scalable, transparent, and reproducible tracking of data use in research literature.

\begin{figure}[H]
    \centering
    \includegraphics[width=1\linewidth]{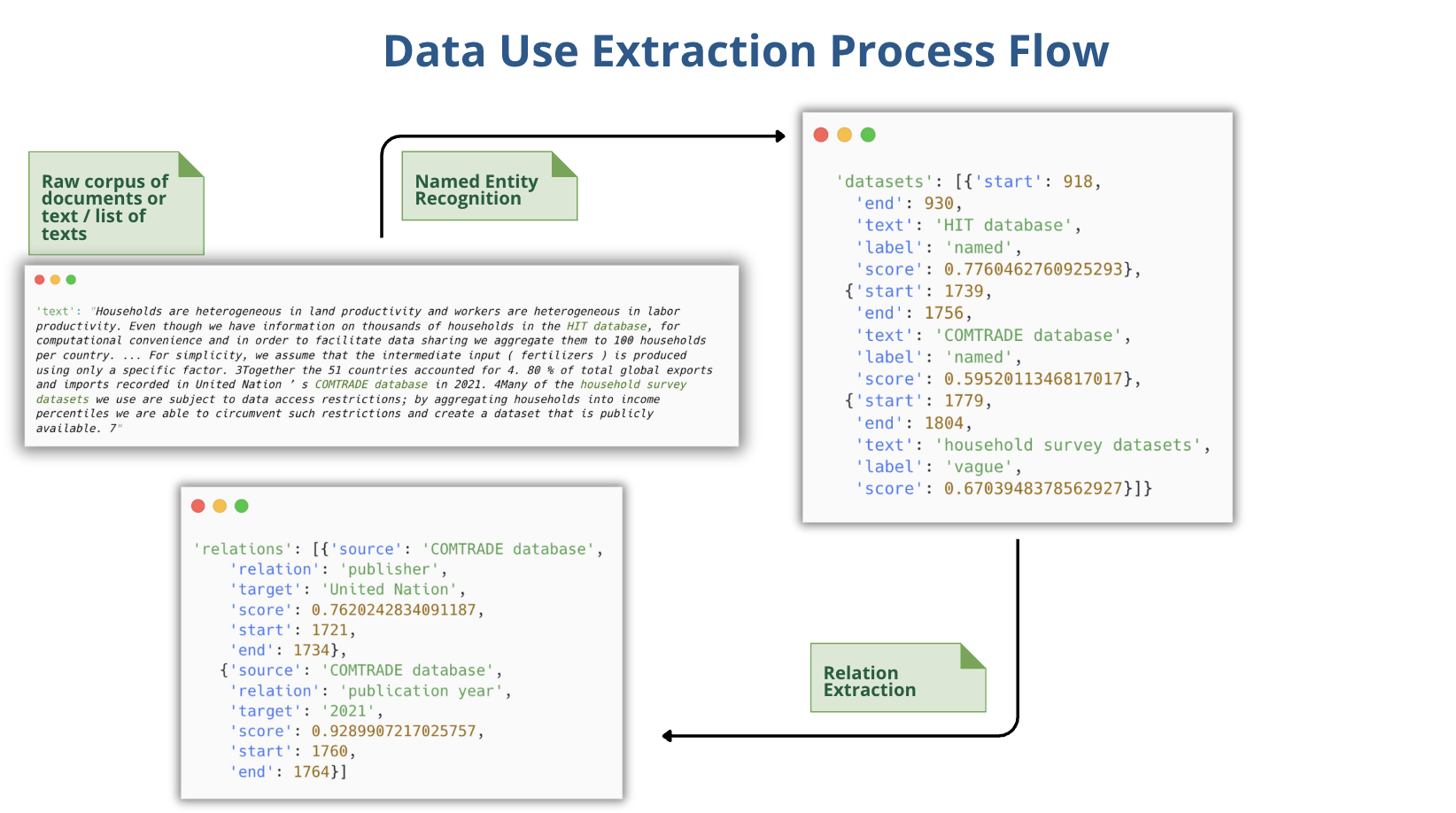}
    \caption{Data-use extraction workflow illustrating the multitask GLiNER-based framework.}
    \label{fig:workflow}
\end{figure}

\section{Methodology}
\label{sec:method}
This section presents the updated methodology for monitoring dataset mentions in research literature. Building on our previous framework, we focus on improving extraction accuracy, interpretability, and efficiency through a unified multitask architecture. To address the limited availability of labeled training data, we continue to employ large-scale synthetic data generation coupled with LLM-based revalidation to construct a weakly supervised training corpus. The overall workflow—covering detecting dataset mention (NER), relation extraction with usage context—is illustrated in Figure~\ref{fig:workflow}.
\subsection{Data}
The primary corpus for model training is derived from the World Bank’s \textit{Policy Research Working Paper} (PRWP) \citep{worldbank2025documents} series.\footnote{\url{https://documents.worldbank.org/en/publication/documents-reports/documentlist?docty_exact=Policy+Research+Working+Paper}}   
We focus on papers addressing topics related to \textit{forced displacement}, \textit{refugees}, and \textit{migration}, as these domains frequently rely on survey, census, and administrative microdata from multiple sources.  
Relevant PRWPs were identified using metadata filters and keyword searches within titles and abstracts (e.g., “refugee,” “displacement,” “asylum,” “migration,” “IDP”).  
Full-text PDFs were downloaded from the open-access repository and converted to plain text using PyMuPDF PDF-to-text extraction.  
Sections containing acknowledgments, references, and tables were excluded to avoid false positives.  
Each processed document was stored with its corresponding metadata—publication year, author, and geographic focus—to support context-aware annotation and evaluation.  
The final filtered subset serves as the foundation for manual annotation and initial model training.
For evaluation, we applied the updated framework to three distinct document sets.  
First, the \textbf{v1 original dataset}—our previously annotated corpus from the initial study—was used to ensure comparability of results across versions.  
Second, we included research and analytical publications from the \textbf{Joint Data Center (JDC) on Forced Displacement}, which provided additional domain coverage related to refugee and displacement studies.  
Finally, we evaluated performance on the broader \textbf{Policy Research Working Paper (PRWP)} collection to assess generalization across topics and writing styles within development research.  
Together, these sources offer a balanced test bed spanning both domain-specific and general research contexts, supporting a robust assessment of the model’s scalability and consistency.

\subsubsection{Manual Annotation using Gradio Interface}

Manual annotation was performed using a custom \texttt{Gradio}-based interface developed to efficiently identify dataset mentions and their associated attributes within full-text research papers (see Appendix~\ref{fig:gradio_example}).  
Annotators highlighted spans of text referring to datasets and categorized them as \texttt{named}, \texttt{unnamed}, or \texttt{vague}.  
They also linked each mention to contextual relations such as \texttt{acronym}, \texttt{publisher}, \texttt{reference population}, \texttt{publication year}, and \texttt{geography}, when available.  
Each annotation record preserved the page number and document source, ensuring traceability and reproducibility of labeled examples.

An example annotation record exported from the Gradio interface is shown below:
\begin{lstlisting}[language=json]
{
  "text": "GBV is particularly prevalent in IDP settlements, among minority clans,
           and in conflict zones ... based on Somalia Demographic and Health Survey (2020).",
  "datasets": [
    {
      "start": 982,
      "end": 1019,
      "text": "Somalia Demographic and Health Survey",
      "label": "named"
    }
  ],
  "relations": [
    {
      "source": "Somalia Demographic and Health Survey",
      "relation": "reference population",
      "target": "Somali women"
    },
    {
      "source": "Somalia Demographic and Health Survey",
      "relation": "publisher",
      "target": "World Bank"
    },
    {
      "source": "Somalia Demographic and Health Survey",
      "relation": "publication year",
      "target": "2020"
    }
  ],
  "source": "Differences-in-Household-Composition-Hidden-Dimensions-of-Poverty-and-Displacement-in-Somalia.pdf",
  "page": 11
}
\end{lstlisting}

Each annotated JSON record captures (i) the raw text segment, (ii) all identified dataset mentions with their offsets and labels, (iii) extracted or annotated relations, and (iv) metadata linking the instance to its document source.  
This structured representation provides a high-fidelity basis for supervised and multitask learning, allowing the model to jointly capture mention boundaries and contextual attributes.

\subsubsection{Synthetic Data Generation}

To expand the diversity and coverage of dataset mentions beyond the forced-displacement domain, we employed large-language-model-based synthetic data generation using the OpenAI GPT API.  
This process addressed the scarcity of labeled examples by creating weakly supervised training data that mimic the linguistic and structural variety found in research literature.  
Prompt templates were designed to condition the model on realistic dataset-usage scenarios and to encourage generation of both named and descriptive dataset mentions.

Each generation round combined seed examples from the manually annotated PRWP corpus with controlled instructions specifying:
\begin{itemize}
    \item the \textbf{type of dataset mention} to generate (named, descriptive, or vague);
    \item the \textbf{domain or topic context} (e.g., health, education, agriculture, environment);
    \item and optional \textbf{metadata fields} such as acronym, producer, year, and geography.
\end{itemize}

The resulting synthetic corpus captured a wide range of phrasing patterns and domain contexts, improving model robustness and generalization during fine-tuning.  
A detailed illustration of the synthetic-data generation workflow, including prompt formulation and integration with annotated data, is provided in Appendix~\ref{fig:synthetic_prompt}.

\subsubsection{Usage Context Data Labeling}

To complement mention and relation data, we automatically generated usage-context labels for each dataset mention in the synthetic corpus.  
This process categorized how each dataset was used within the surrounding text as \textbf{primary}, \textbf{supporting}, or \textbf{background}.  

We employed an LLM-based classification script that provided the model with the document text, extracted dataset mentions, and explicit role definitions.  
The model was prompted to output structured JSON containing the mention, its assigned label, and a short rationale explaining the classification.  
This approach enabled rapid creation of labeled examples without manual annotation while maintaining interpretability through natural-language justifications.

An example of the core logic is shown below, where each entry is passed to the LLM and parses the structured response for integration into the synthetic dataset:

\begin{lstlisting}[language=json]
You are an expert research assistant. Classify the role of EACH dataset mention in the text.

Definitions:
- Background: Mentioned as prior/related work, not analyzed in the study.
- Supporting: Complements findings but not central to analysis.
- Primary: Main dataset analyzed in the study.

Expected JSON Output:
{
  "results": [
    {
      "mention": "<exact mention>",
      "label": "<Background | Supporting | Primary>",
      "explanation": "<1-3 sentence rationale>"
    }
  ]
}
\end{lstlisting}

This automated labeling pipeline substantially reduced manual effort while providing high-quality supervision signals for fine-tuning the multitask GLiNER model.

\subsubsection{LLM-as-a-JUDGE Revalidation Pipeline}

To ensure label consistency and reduce false positives in the expanded dataset, we introduced an LLM-based revalidation pipeline.  
This stage acts as an automated quality-control filter between model inference and retraining.  
Each extracted record—comprising \texttt{text}, \texttt{datasets}, and \texttt{relations}—is re-evaluated by the LLM using a structured validation prompt (see Appendix A.3, Figure A.3).

The revalidation prompt defines what constitutes a valid dataset mention and relation, distinguishing between \textit{named}, \textit{descriptive}, and \textit{vague} datasets.  
For each predicted mention, the model must decide whether to:
\begin{itemize}
    \item \texttt{keep} the mention if it refers to a valid dataset or data collection;
    \item \texttt{remove} it if it corresponds to a non-dataset entity such as organizations, institutions, or conceptual frameworks.
\end{itemize}
Relations are then cross-checked to ensure that each link corresponds to a retained dataset mention.  
The model outputs the same JSON structure, augmented with an \texttt{action} (keep/remove) and a \texttt{reason} field for interpretability.

This validation loop leverages the LLM’s contextual reasoning to enforce the annotation schema without requiring additional human supervision.  
By filtering inconsistent or spurious predictions before retraining, the pipeline improves data precision and stabilizes subsequent fine-tuning performance.

A complete example of the revalidation prompt is provided in Appendix~\ref{fig:revalidation_prompt}, including instructions and structured examples used during inference.

\subsection{Multitask GLiNER Architecture}

To extend our previous work on dataset-mention extraction, we adopt a multitask variant of the Generalist Language Inference for Named Entity Recognition (GLiNER) model \citep{zaratianna2023gliner}.  
This version integrates three interrelated tasks—dataset mention extraction, relation identification, and usage-context classification—within a single transformer-based framework.

The multitask configuration equips the model with three integrated capabilities central to dataset-use monitoring.  
First, it identifies dataset mentions—both formally named resources and descriptive references to data collections—within the text.  
Second, it extracts associated metadata and contextual attributes such as producer, acronym, publication year, data geography, and reference population.  
Finally, it interprets how each dataset is used in the study by classifying its contextual role as primary, supporting, or background.  
Together, these capabilities enable a structured and interpretable representation of dataset usage derived directly from unstructured text.

This unified design allows the model to capture dependencies between dataset mentions, their descriptive attributes, and their functional roles within the surrounding narrative.  
By training all components jointly, the framework produces more coherent and traceable structured outputs—supporting scalable monitoring of dataset usage across research literature.

\subsubsection{NER Component}

The named entity recognition (NER) component serves as the foundation of the framework, identifying spans of text that correspond to datasets or data collections mentioned within research papers.  
Its goal is to detect a wide range of dataset references, from explicitly named resources such as \textit{Demographic and Health Surveys (DHS)} or \textit{Living Standards Measurement Study (LSMS)}, to more descriptive mentions like “household income survey data” or “administrative education records.”

The model formulates dataset mention detection as a token-level classification task.  
Using a transformer encoder, it assigns labels to each token under a BIO-style tagging scheme and reconstructs contiguous spans representing dataset mentions.  
This token-based approach allows the model to accurately locate dataset names within complex sentence structures and handle diverse citation styles, abbreviations, and punctuation patterns commonly found in research writing.  
Unlike rule-based or template-driven methods, it generalizes effectively across domains and writing conventions without requiring handcrafted linguistic patterns.

Each identified mention is assigned one of three labels according to its specificity and clarity:
\begin{itemize}
    \item \textbf{Named} – formally cited datasets or surveys with recognizable titles or acronyms;
    \item \textbf{Descriptive} – descriptive references to data collections without explicit names;
    \item \textbf{Vague} – indirect or ambiguous mentions of data, e.g., “administrative data”
\end{itemize}

Accurate mention detection is crucial for downstream processing.  
It anchors all subsequent stages—relation extraction and usage-context classification—ensuring that the model builds a coherent structured representation of dataset usage grounded in precise textual evidence.

\subsubsection{Relation Extraction Component}

The relation extraction component identifies and links additional information associated with each detected dataset mention.  
For every dataset span recognized by the NER component, the model extracts structured attributes that describe its key metadata and contextual properties.  
These include the \textbf{producer or publisher}, \textbf{author or creator}, \textbf{acronym}, \textbf{data description}, \textbf{data type}, \textbf{data geography}, \textbf{publication year}, \textbf{reference year}, and \textbf{reference population}, as well as the \textbf{usage context} describing how the dataset is utilized within the paper (e.g., primary, supporting, or background).  
By capturing this full relational schema, the model constructs a comprehensive representation of dataset usage that connects each mention to its semantic and bibliographic context.

Relations are modeled at the token level within the same encoder, allowing the system to learn co-occurrence and syntactic cues that signal attribute–dataset associations.  
For instance, phrases such as “collected by the World Bank” or “2017–18 Kenya DHS” provide strong textual evidence linking an entity to its producer or year.  
By representing relations in this way, the model can capture both explicit metadata (e.g., “Demographic and Health Surveys (DHS), 2014”) and implicit ones embedded within complex clauses (e.g., “data from surveys conducted in Kenya”).  

This component is essential for transforming raw mention detection into a structured metadata representation.  
It enables downstream analysis of how datasets are described, produced, and referenced across studies—offering a clearer view of the provenance and reuse of data within the research literature.  
Each extracted triple is stored in the format \texttt{(source, relation, target)}, which supports consistent evaluation and easy integration into structured databases or visualization tools.

\subsubsection{Usage Context Classification}\label{sec:usagecontextlabeling}

The usage-context classification component determines the functional role that each dataset mention plays within a research paper.  
While the NER and relation components focus on identifying datasets and their metadata, this component interprets \textit{how} the dataset is used in context—distinguishing between primary data analyzed in the study and those cited for reference or comparison.

Each mention is classified into one of three roles:
\begin{itemize}
    \item \textbf{Primary} – the dataset serves as the main source of empirical analysis (e.g., “We use the 2017–18 Somalia High-Frequency Survey to examine...”);
    \item \textbf{Supporting} – the dataset is referenced to validate or complement findings (e.g., “We compare our results with those from the DHS data...”);
    \item \textbf{Background} – the dataset is mentioned only as prior work or contextual information (e.g., “Previous studies have used the LSMS dataset...”).
\end{itemize}

Usage-context data were generated using an LLM-assisted annotation process (Section~\ref{sec:usagecontextlabeling}), where each dataset mention was automatically assigned a role label with a short textual rationale.  
These labeled examples were then incorporated into model training, enabling the system to learn linguistic cues—such as verbs of analysis, comparison, or citation—that correlate with different dataset roles.

This component enriches the framework by moving beyond surface-level recognition toward semantic interpretation of data use.  
By linking each dataset mention to its contextual role, the model provides a structured view of not only \textit{what} data are used but also \textit{how} they are used across research literature.

\subsection{Training and Optimization}

\textbf{Fine-tuning data}  
The multitask GLiNER model was fine-tuned on a composite dataset comprising (i) manually annotated samples from the PRWP corpus,  
(ii) synthetically generated examples that expanded coverage across multiple domains, and  
(iii) revalidated outputs filtered through the LLM-based quality control pipeline.  
The final training set integrated these three sources to balance precision from manual labels with the diversity and scale of synthetic data.  
A small held-out portion of the annotated corpus, along with the v1 original dataset and JDC documents, was reserved for validation to monitor overfitting and assess generalization performance.

\textbf{Training setup}  
We used the official \texttt{gliner} training scripts\footnote{\url{https://github.com/urchade/GLiNER/blob/main/train.py}} as the base implementation, modifying it to support multitask learning.  
The customization enabled joint optimization of three supervised objectives—dataset mention extraction, relation extraction, and usage-context classification—under a shared encoder.
This adaptation allowed the model to learn interdependent features between mention spans, metadata relations, and contextual roles without requiring separate training runs.

\textbf{Configuration.}  
Training followed a two-stage fine-tuning process.  
In the first stage, the model was \textit{pre-fine-tuned} on the synthetic dataset to establish general familiarity with dataset-related language patterns across multiple domains.  
This stage used a cosine learning-rate scheduler with a relatively higher learning rate ($5\times10^{-5}$) and smaller batch size (4), allowing the model to explore a wide parameter space while avoiding overfitting on any single context.  
The pre-fine-tuning objective prioritized broad recall and robustness to linguistic variation in how datasets are mentioned.  

In the second stage, the same model was \textit{fine-tuned} on the curated, high-quality validated dataset containing manually annotated and LLM-revalidated examples.  
This stage applied a lower learning rate ($5\times10^{-6}$), linear decay scheduler, and larger batch size (8) to refine the model’s predictions toward higher precision and domain-specific consistency.  
The fine-tuning objective focused on alignment with the dataset-mention schema and accurate relation and usage-context extraction.  

Both stages used focal loss with $\alpha = 0.75$ and $\gamma = 2$ to handle class imbalance among mention types and relation labels.  
Weight decay was fixed at $0.01$ for both, with a $10\%$ warm-up ratio.  
Training was conducted using the Hugging Face \texttt{Trainer} API and customized versions of the official \texttt{gliner} training scripts to support multitask optimization across NER, relation extraction, and usage-context classification tasks.

All experiments were executed on an NVIDIA A100 GPU (80~GB), with mixed-precision training enabled.  
The two-stage procedure proved effective in stabilizing convergence and balancing generalization (from synthetic data) with precision (from validated data).

\section{Results}
Our evaluation assesses the performance of dataset extraction models for identifying and classifying dataset mentions in research papers.  
Table~\ref{tab:comparison} summarizes the performance of the models across different training configurations and baselines.  
In addition to evaluating the original (v1) models from the previous study, we re-evaluated those baselines on the same annotated corpus and extended the analysis to include two granular levels of evaluation:  
\textbf{(i)} passage-level evaluation, which measures model performance per text segment, and  
\textbf{(ii)} document-level evaluation, which aggregates predictions across entire papers.  
This dual setup provides a more comprehensive view of model precision and consistency when applied to larger document contexts.

\begin{table}[H]
\centering
\caption{Performance of Extraction Models for Data Use}
\label{tab:comparison}
\begin{tabular}{lccc}
\toprule
\multicolumn{4}{c}{\textbf{Data Use Extraction Models}} \\
\midrule
\textbf{Model [Training data]} & \textbf{Precision} & \textbf{Recall} & \textbf{F$_\beta$-score} \\
\midrule
Phi-3-mini [Synthetic and curated] & 69.45 & 80.65 & 71.43 \\
Phi-3-mini [Synthetic only] & 60.00 & 70.00 & 61.76 \\
Phi-3-mini [Curated only] & 55.68 & 65.52 & 57.58 \\
GLiNER-large-v2.1 & 62.50 & 71.43 & 64.10 \\
NuExtract-v1.5 & 20.97 & 46.43 & 23.55 \\
\midrule
\multicolumn{4}{c}{\textbf{Fine-Tuned Multitask GLiNER (Our v2)}} \\
\midrule
\textbf{v1-orig [passage]} & \textbf{1.00} & \textbf{0.81} & \textbf{0.90} \\
\textbf{v1-orig [document]} & \textbf{1.00} & \textbf{0.88} & \textbf{0.94} \\
\textbf{JDC [passage]} & \textbf{0.99} & \textbf{0.92} & \textbf{0.95} \\
\textbf{JDC [document]} & \textbf{0.99} & \textbf{0.93} & \textbf{0.96} \\
\textbf{PRWP [passage]} & \textbf{0.99} & \textbf{0.83} & \textbf{0.90} \\
\textbf{PRWP [document]} & \textbf{0.99} & \textbf{0.87} & \textbf{0.92} \\
\bottomrule
\end{tabular}
\end{table}

\begin{equation}
J(S_1, S_2) = 
\frac{|W_1 \cap W_2|}
{|W_1| + |W_2| - |W_1 \cap W_2|}
\end{equation}

In Equation~(1), $J(S_1, S_2)$ represents the Jaccard similarity between two strings $S_1$ and $S_2$.  
The sets $W_1$ and $W_2$ contain the unique tokens derived from each string.  
The numerator $|W_1 \cap W_2|$ denotes the number of overlapping tokens, while the denominator 
$|W_1| + |W_2| - |W_1 \cap W_2|$ ensures that shared words are not double-counted, producing a similarity score between 0 and 1.  
A Jaccard score greater than 0.5 is considered a match.  
Based on this classification, precision, recall, and F$_\beta$-scores are computed to evaluate extraction performance.

Our results show that the fine-tuned multitask GLiNER model consistently outperforms prior baselines across all evaluation datasets and granularities.  
Compared to the earlier two-stage fine-tuning framework, the unified multitask setup demonstrates superior stability and generalization when applied to both passage- and document-level extractions, outperforming open benchmark systems such as \textit{GLiNER-large-v2.1} and \textit{NuExtract} \citep{cripwell2024nuextract}.

\begin{itemize}
    \item On the original (v1) dataset, the multitask model attains near-perfect precision ($1.00$) and robust recall (0.81 at the passage level, 0.88 at the document level), indicating strong fidelity in reproducing valid dataset mentions while capturing additional instances missed by earlier models.
    \item Across the JDC corpus, performance remains consistently high (F$_\beta$ scores of 0.95–0.96), showing that the model generalizes effectively to new document collections from similar domains without domain-specific fine-tuning.
    \item Evaluation on PRWP papers yields slightly lower recall (0.83–0.87) yet maintains precision above 0.99, suggesting that while the model avoids false positives even in heterogeneous text, additional exposure to new writing styles and citation formats could further improve sensitivity.
\end{itemize}

These results confirm that the multitask integration of entity, relation, and context extraction substantially enhances end-to-end consistency compared to the prior pipeline.  
The unified model benefits from shared contextual representations across tasks, which reduces error propagation and enables better structural alignment between dataset mentions and their attributes.  

The results also highlight the effect of aggregation granularity: document-level evaluation consistently improves recall and overall F$_\beta$ scores compared to passage-level metrics, reflecting the model’s capacity to capture dispersed dataset references across paragraphs within the same paper.  
This improvement underscores the advantage of viewing dataset mention monitoring as a document-level problem rather than a sentence-level classification task.

In summary, the finetuned multitask GLiNER framework delivers significant gains in precision and recall while maintaining interpretability and generalizability across corpora.  
Its ability to jointly reason over entity spans, relations, and usage contexts establishes a scalable foundation for reliable dataset-use monitoring across diverse research literature.

\subsection{Discussion}

Our findings demonstrate that combining dataset mention extraction, relation identification, and usage-context classification within a single modeling framework improves the overall reliability and interpretability of dataset-use monitoring.  
Compared to the previous two-stage fine-tuning framework, the multitask GLiNER model achieves higher precision and recall across all evaluation sets—reaching F$_\beta$ scores of up to 0.96 on document-level evaluation—while maintaining consistent performance across distinct corpora such as PRWP and JDC.  

The model’s improved precision indicates a stronger ability to isolate valid dataset mentions, while its high recall highlights better generalization to unseen phrasing and domain variations.  
These gains are largely attributed to the combination of synthetic data generation and LLM-based revalidation, which together provide a high-quality training signal while minimizing noise from weakly labeled examples.  
The revalidation stage, in particular, plays a critical role in filtering out false positives and enforcing labeling consistency, contributing to the model’s stability across different validation sets.  

Finally, our comparison of passage- and document-level evaluations shows that aggregating predictions across entire documents yields more complete representations of dataset usage, as datasets are often mentioned in multiple sections of a paper.  
This finding supports evaluating dataset monitoring methods at the document level to better reflect real-world data citation patterns and to generate more policy-relevant insights into data reuse and accessibility.

\section{Conclusion}

This paper presented an updated methodology for monitoring dataset mentions in research literature, introducing a multitask GLiNER framework that unifies mention extraction, relation identification, and usage-context classification into a single model.  
Building on our earlier approach that leveraged synthetic data and LLM-based revalidation, this new framework simplifies training while achieving state-of-the-art performance in structured dataset metadata extraction.

By jointly optimizing related tasks, the model achieves consistent gains in precision, recall, and interpretability across different corpora, demonstrating the effectiveness of multitask learning for structured information extraction in low-resource settings.  
The combination of synthetic data generation, curated fine-tuning, and post-hoc revalidation provides a scalable foundation for extending dataset-use monitoring across disciplines.

Future work will focus on expanding annotated corpora beyond current domains, improving canonicalization of extracted datasets, and exploring adaptive fine-tuning strategies for domain transfer.  
We also plan to integrate temporal tracking of dataset mentions to support longitudinal analyses of data reuse and citation practices.  
Ultimately, this line of work contributes toward solving the broader challenge of arbitrary dataset citation—advancing transparency, discoverability, and accountability in the use of research data.

\subsubsection*{Acknowledgments}
This work is supported by the “KCP IV - Exploring Data Use in the Development Economics
Literature using Large Language Models (AI and LLMs)” project funded by the Knowledge
for Change Program (KCP) of the World Bank - RA-P503405-RESE-TF0C3444.

\subsubsection*{Disclaimer and Disclosure of AI Use}
The findings, interpretations, and conclusions expressed in this paper are entirely those of
the authors. They do not necessarily represent the views of the International Bank for
Reconstruction and Development/World Bank and its affiliated organizations, or those of
the Executive Directors of the World Bank or the governments they represent.
This work used AI tools at various stages, including generating synthetic data and reasoning
using the gpt-5-mini model (API) and open source AI models. In addition, Microsoft CoPilot and ChatGPT were employed to enhance the manuscript’s readability.

\bibliography{iclr2025_conference}
\bibliographystyle{iclr2025_conference}

\appendix
\section{Appendix}
A. Data details, B. Annotation schema, C. Prompt templates, D. Manual annotation interface, E. Metric definitions, F. Additional results.

\subsection{A. Data Sources}
This section provides additional details on the datasets used for training and evaluation.
\begin{itemize}
    \item \textbf{World Bank Documents and Reports} – repository of open-access project, policy, and research documents, filtered by forced displacement and refugee-related keywords.
    \item \textbf{Policy Research Working Papers (PRWP)} – selected open-access working papers focusing on socioeconomic development, displacement, and climate-related themes.
    \item \textbf{Joint Data Center (JDC) Corpus} – a curated set of analytical and research papers related to forced displacement.
\end{itemize}

\vspace{-0.5em}

\subsection{B. Annotation Schema and Guidelines}

Table~\ref{tab:annotation-schema} presents the complete entity and relation schema used for annotation and model training.  
Entities represent dataset mentions within the text, while relations link each mention to its associated metadata attributes such as producer, year, or geography.  
These labels are used consistently across both manually annotated and synthetic datasets.

\begin{table}[!htbp]
\centering
\caption{Annotation schema for dataset mentions and relations.}
\label{tab:annotation-schema}
\begin{tabular}{ll}
\hline
\textbf{Entity Type} & \textbf{Description} \\
\hline
\texttt{named} & Explicitly cited datasets with formal titles (e.g., DHS, LSMS). \\
\texttt{unnamed} & Descriptive dataset mentions (e.g., “household income survey”). \\
\texttt{vague} & Ambiguous or indirect dataset mentions (e.g., “the collected data”). \\
\hline
\textbf{Relation Type} & \textbf{Description} \\
\hline
\texttt{acronym} & Short-form identifier or abbreviation for the dataset (e.g., DHS, LSMS). \\
\texttt{author} & Creator, compiler, or principal investigator of the dataset. \\
\texttt{data description} & Thematic scope or description of the dataset (e.g., household, firm-level). \\
\texttt{data geography} & Geographic coverage or spatial region represented by the dataset. \\
\texttt{data type} & Type or modality of the data (e.g., survey, administrative, geospatial). \\
\texttt{publication year} & Year when the dataset was released or published. \\
\texttt{publisher} & Institution or organization responsible for publishing or maintaining the dataset. \\
\texttt{reference population} & Population or subgroup represented in the dataset (e.g., households, refugees). \\
\texttt{reference year} & Reference or collection period covered by the data. \\
\texttt{usage context} & Functional role of the dataset in the paper (primary, supporting, background). \\
\hline
\end{tabular}
\end{table}

\vspace{-0.5em}
\noindent\textbf{Usage Context Definitions.}  
Each dataset mention is assigned one of three contextual roles:
\begin{itemize}
    \item \textbf{Primary} — The dataset serves as the main source of analysis or evidence for the study’s findings.
    \item \textbf{Supporting} — The dataset is used to validate, compare, or complement other data sources.
    \item \textbf{Background} — The dataset is referenced contextually or as part of related literature without being analyzed directly.
\end{itemize}

\vspace{-0.5em}

\subsection{C. Synthetic Data Generation and Revalidation Prompts}

To support low-resource dataset mention extraction, two types of LLM prompts were designed:  
(1) for generating synthetic dataset-mention examples across diverse research contexts, and  
(2) for revalidating model predictions to filter out invalid or incorrectly labeled mentions.  
Both prompt templates are illustrated below for reproducibility.

\vspace{-0.5em}
\begin{figure}[!htbp]
\centering
\includegraphics[width=0.95\linewidth]{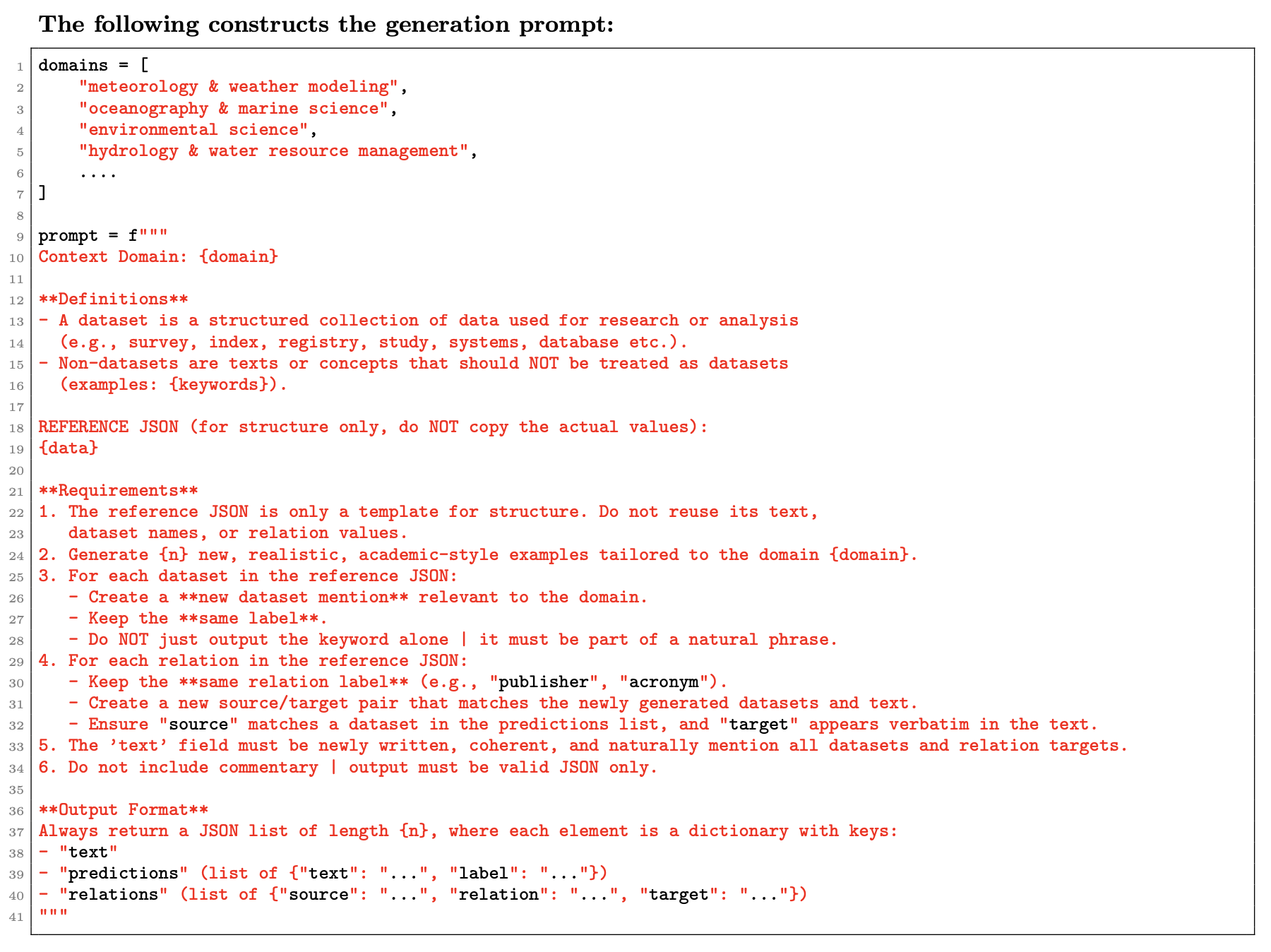}
\caption{Prompt template used for synthetic data generation. The prompt guides the LLM to produce controlled examples of dataset mentions across varied research domains and sentence structures.}
\label{fig:synthetic_prompt}
\end{figure}

\vspace{-1em}
\begin{figure}[!htbp]
\centering
\includegraphics[width=0.95\linewidth]{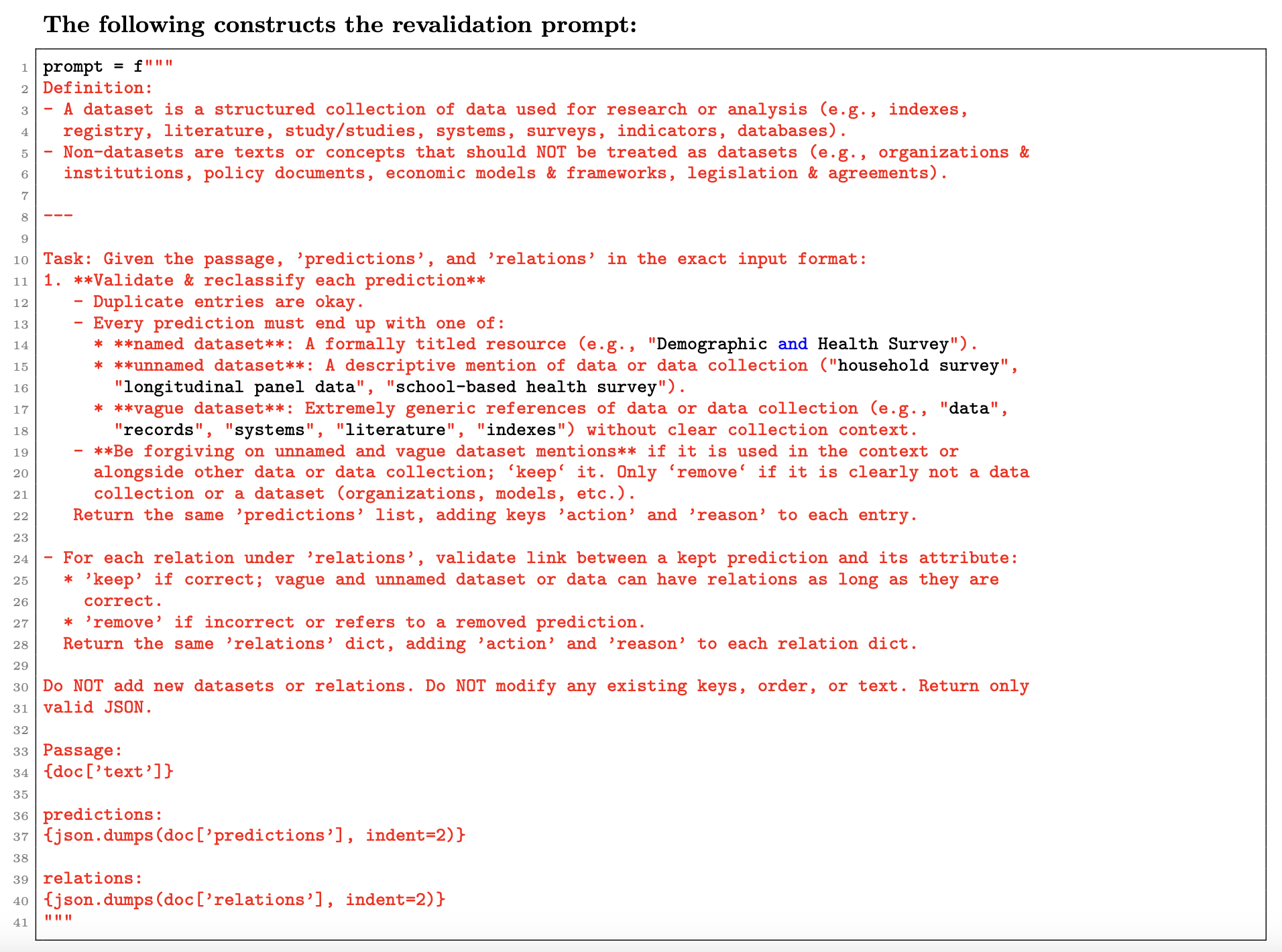}
\caption{Prompt template used for LLM-based revalidation. This ensures that only valid dataset mentions—consistent with labeling criteria—are retained in the training corpus.}
\label{fig:revalidation_prompt}
\end{figure}

\vspace{-0.5em}

\subsection{D. Manual Annotation Interface}

To create high-quality labeled data for fine-tuning, we developed a custom Gradio-based web application hosted on Hugging Face Spaces.  
The interface allows annotators to highlight dataset mentions directly within research text, assign entity types (\texttt{named}, \texttt{unnamed}, or \texttt{vague}), and define relations such as publisher, geography, and year.  
Annotations are exported in structured JSON format for downstream processing.

\vspace{-0.5em}
\begin{figure}[!htbp]
\centering
\includegraphics[width=0.95\linewidth]{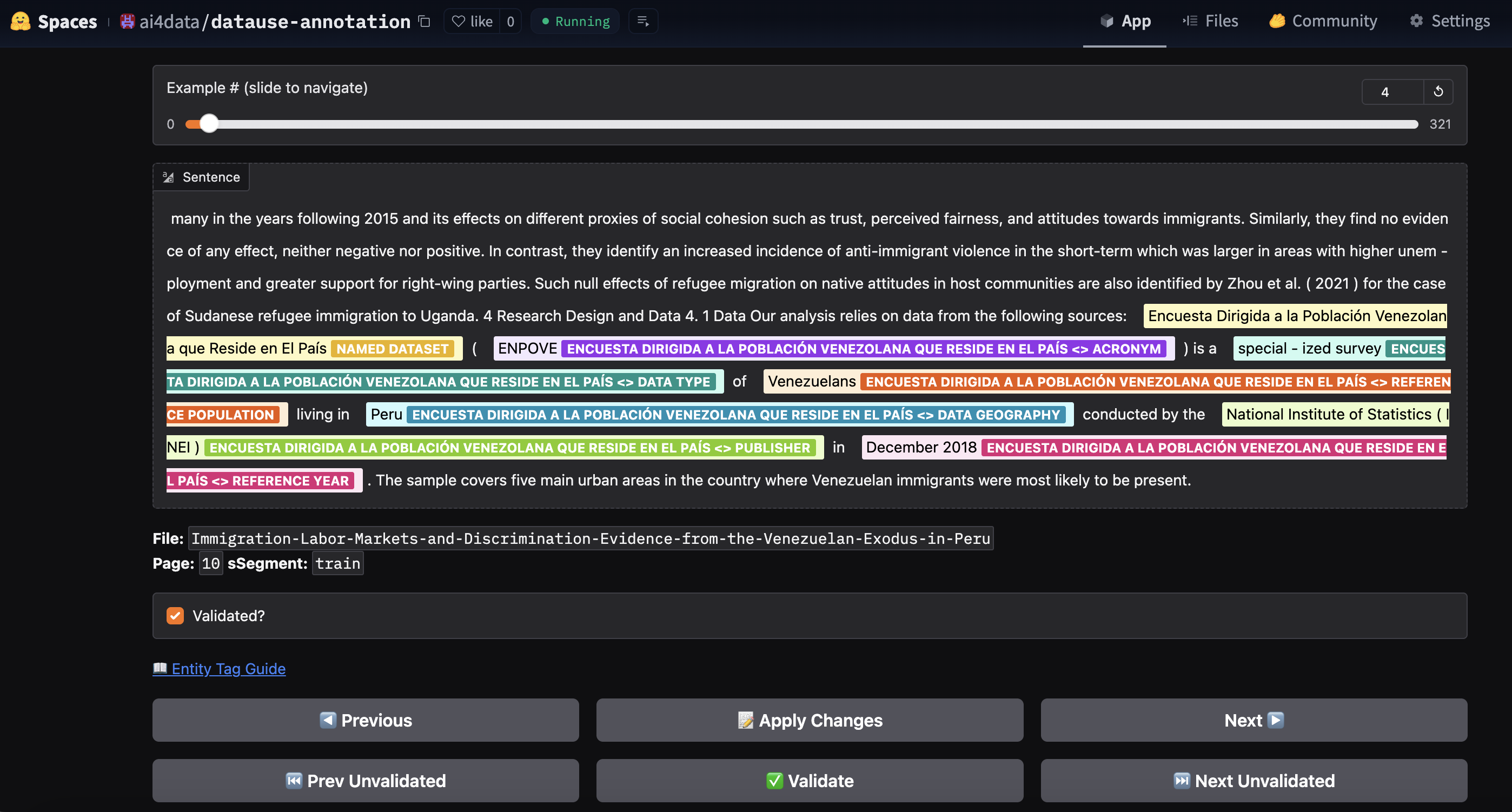}
\caption{Gradio annotation interface hosted on Hugging Face Spaces, enabling manual labeling of dataset mentions and relations in research papers.}
\label{fig:gradio_example}
\end{figure}

\vspace{-0.5em}

\subsection{E. Metric Definition}
For evaluation, we adopt the Jaccard-based F$_\beta$ score used in the prior work \citep{solatorio2025ai}:
\[
J(S_1, S_2) = \frac{|W_1 \cap W_2|}{|W_1| + |W_2| - |W_1 \cap W_2|}
\]
where $W_1$ and $W_2$ are token sets. A Jaccard score $\geq 0.5$ is considered a match.

\vspace{-0.5em}

\subsection{F. Additional Results}
Extended ablation results, visualizations, and performance breakdowns across document domains (PRWP, JDC, and v1-original) are available in the project repository:
\url{https://github.com/worldbank/ai4data}.

\end{document}